\documentclass[letterpaper, 10 pt, conference]{ieeeconf}  

\IEEEoverridecommandlockouts                              

\usepackage{bm}
\usepackage[utf8]{inputenc}
\usepackage[T1]{fontenc}

\usepackage[english]{babel}                     
\usepackage{amsmath, amssymb}           
\usepackage{float}                              
\usepackage[final, colorlinks = true]{hyperref} 
\usepackage{graphics}                           
\usepackage{epsfig}                             
\usepackage{microtype}                          
\usepackage{subfig}
\usepackage{booktabs}
\usepackage{csquotes}                           
\usepackage[yyyymmdd]{datetime}                 
\usepackage{pseudocode}
\usepackage[ruled]{algorithm2e}
\usepackage{cite}

\newcommand*{\argmin}{\operatornamewithlimits{argmin}}

\newcommand*{\set}[1]{{\mathcal{\MakeUppercase{#1}}}}			

\newcommand*{\seq}[1]{\MakeUppercase{#1}}

\newcommand*{\mean}[1]{\overline{#1}}

\renewcommand{\vec}[1]{{\boldsymbol{\mathbf{#1}}}}
\newcommand*{\mat}[1]{{\MakeUppercase{#1}}}
\newcommand*{\transpose}{\mathsf{T}}

\newcommand*{\simparams}{\vec{\theta}} 

\newcommand*{\trajectory}{\vec{\tau}}

\newcommand*{\normal}{\mathcal{N}}					

\newcommand*{\diff}{{\mathop{}\operatorname{d}}}

\newcommand*{\weight}{\omega}
\newcommand*{\normaliser}{\eta}

\newcommand*{\state}{\vec{x}}

\newcommand*{\stateDim}{n}

\newcommand*{\seqOp}{\mathcal{H}}

\newcommand*{\stimuli}{\vec{v}}

\newcommand*{\stimSeq}{V}

\newcommand*{\control}{\vec{u}}
\newcommand*{\controlSeq}{\seq{u}}
\newcommand*{\controlSpace}{\set{U}}
\newcommand*{\controlDim}{m}

\newcommand*{\instCost}{c}
\newcommand*{\termCost}{\phi}

\newcommand*{\stimCost}{S}
\newcommand*{\Loss}{\mathcal{L}}
\newcommand*{\invTemperature}{\lambda}
\newcommand*{\controlMin}{\beta}
\newcommand*{\controlAuth}{\mat{\Sigma}}
\newcommand*{\controlHorizon}{T}
\newcommand*{\controlSamples}{K}

\newcommand*{\sigmaParam}{\vec{\chi}}
\newcommand*{\sigmaState}{\vec{\gamma}}
\newcommand*{\primScaling}{\nu}
\newcommand*{\secScaling}{\kappa}
\newcommand*{\spread}{\alpha}
\newcommand*{\degree}{\xi}
\newcommand*{\sigmaWeight}{\varpi}
\newcommand*{\sigmaLength}{L}

\newcommand*{\xicr}{x_{{\operatorname{ICR}}}}
\newcommand*{\wradius}{r_{{\operatorname{w}}}}
\newcommand*{\axdist}{a_{{\operatorname{w}}}}

\newcommand*{\covMatrix}{\mat{\Sigma}}

\newcommand*{\pluseq}{\mathrel{+}=}

\newcommand*{\iid}{i.i.d.\xspace}
\newcommand*{\wrt}{w.r.t.\xspace}

\title{\LARGE \bf
DISCO: Double Likelihood-free Inference Stochastic Control
}

\author{Lucas Barcelos$^{*,1}$, Rafael Oliveira$^{1}$, Rafael Possas$^{1}$, Lionel Ott$^{1}$, and Fabio Ramos$^{1, 2}$
\thanks{$^*$ Corresponding author: lucas.barcelos@sydney.edu.au}
\thanks{$^1$ School of Computer Science, The University of Sydney, Australia}
\thanks{$^2$ NVIDIA, USA}%
\thanks{Code available at: https://github.com/lubaroli/disco}
}

\begin{document}

\maketitle
\thispagestyle{empty}
\pagestyle{empty}

\begin{abstract}
Accurate simulation of complex physical systems enables the development, testing, and certification of control strategies before they are deployed into the real systems. As simulators become more advanced, the analytical tractability of the differential equations and associated numerical solvers incorporated in the simulations diminishes, making them difficult to analyse. A potential solution is the use of probabilistic inference to assess the uncertainty of the simulation parameters given real observations of the system. Unfortunately the likelihood function required for inference is generally expensive to compute or totally intractable. In this paper we propose to leverage the power of modern simulators and recent techniques in Bayesian statistics for likelihood-free inference to design a control framework that is efficient and robust with respect to the uncertainty over simulation parameters. The posterior distribution over simulation parameters is propagated through a potentially non-analytical model of the system with the unscented transform, and a variant of the information theoretical model predictive control. This approach provides a more efficient way to evaluate trajectory roll outs than Monte Carlo sampling, reducing the online computation burden. Experiments show that the controller proposed attained superior performance and robustness on classical control and robotics tasks when compared to models not accounting for the uncertainty over model parameters.  

\end{abstract}

\section{Introduction}
Robustness to model miss-specification and noisy sensor measurements is a critical property for control systems operating in complex robotics applications. The development of powerful and more realistic simulators allows practitioners to analyse and verify the performance of the controller against these variables before the controller is deployed to the real robot. In Model Predictive Control (MPC) one seeks to iteratively find the solution of an optimisation problem for a receding finite time-horizon using an approximate model of the system. When the dynamic model is given by complex simulators that incorporate differential equations and numerical solvers there is little hope the equations can be reversed to reason about the parameters of the simulation to best match the real behaviour of the system. Furthermore, the simulator might abstract away the equations and solver from the user. Effectively, it can be interpreted as a generative model that can be sampled from given a set of parameter values, but not inverted. In this paper we pose the question, can we leverage the power of simulators, treated as generative models, to design control strategies that are robust to parameter uncertainty?

On the other hand, the application of MPC to linear systems has been an active research area for many decades with extensive  deployments to many practical problems \cite{de_oliveira_multi-agent_2010,mayne_model_2014, mesbah_stochastic_2016}.
Notably, the most common setting for linear MPC application are tasks that involve trajectory tracking or stabilisation.
However, control tasks in reinforcement learning are usually more complex and therefore less suitable to linearisation, motivating the use of non-linear models \cite{recht_tour_2018}.
Another motivation for more complex models is the ability to use of more expressive constraints, even if not directly involved in the physical process, such as economic criteria \cite{bradford_stochastic_2017}.
Despite its vast application in the linear case, the use of MPC in non-linear systems continues to be an increasingly active area of research in control theory \cite{mayne_model_2014, mesbah_stochastic_2016}.

Recent work in the field has led to controllers that are able to incorporate non-linear dynamics without relying on linear or quadratic approximations \cite{williams_information_2017, williams_information-theoretic_2018}. 
However, most MPC controllers still do not consider uncertainty in the parameters of their internal simulator for future trajectories. In addition, estimating parameters for the system's model usually requires large amounts of data from the real system, which can be infeasible for some applications.
Yet, whenever the stochastic system uncertainties can be adequately modelled, it is more natural to explicitly take them into account in the control design method itself. In Stochastic MPC, the uncertainty on the internal system dynamics is intrinsic to the optimal control problem solved at every time step. This allows the controller to trade-off performance and satisfaction of the constraints by regulating the joint probability distribution of the system states and outputs \cite{mesbah_stochastic_2016}.

In this paper we make the following contributions: we develop a Stochastic Non-linear MPC variant which leverages recent advancements in likelihood-free inference to estimate both the uncertainty on the simulator parameters as well as to propagate it throughout the estimated trajectories. We call our method double likelihood-free inference stochastic control, DISCO.
The posterior distribution for the parameters of the simulator is estimated by combining simulated data from generative models and observations from the physical system. Using this posterior distribution allows us to take into account the uncertainty about the system's dynamics in the decision making process during the control task.
We proceed to show that the Unscented Transform (UT) \cite{julier_unscented_2004} provides a computationally efficient alternative when compared to traditional Monte Carlo approaches to propagate the uncertainty from the parameter space to the forward modelling of the trajectory roll outs.
In short, DISCO can be seen as a variant of the Information Theoretical MPC (IT-MPC) control algorithm \cite{williams_information-theoretic_2018} that considers the uncertainty in the system's parameters in its internal trajectory simulations.

\section{Related work}
The use of MPC in the control of linear systems is very mature and has been widely studied and applied to real systems.
However, as seen in \cite{mayne_model_2014}, Non-linear MPC (NMPC) is still an open-research question, especially for systems were uncertainty over parameters and constraints on controls and state-space are considered.
The most common methods for controlling general nonlinear systems are based on Non-linear Programming \cite{houska_acado_2011} and Differential Dynamic Programming (DDP) \cite{erez_integrated_2013}. 
Both rely on approximations of dynamics and cost functions so that the online optimisation problem becomes tractable.
However, these mainstream gradient-based MPC approaches have some shortcomings.
In the DDP method, the cost function must be smooth and it is notoriously difficult to include state constraints.
Whereas with nonlinear programming constraints may be easily accounted for, but a common issue is what to do when no feasible solution is found.

In \cite{mesbah_stochastic_2016}, a family of Stochastic NMPC (SNMPC) methods are discussed.
In Tube-based NMPC the objective of the control policy is to ensure that the forward trajectories will remain inside a desirable tube centred around a given trajectory, however the boundary tube has to be computed offline \cite{rakovic_parameterized_2012}.
A multi-stage NMPC approach has been suggested in which the uncertainty is modelled by a scenario tree approach from stochastic programming.
However, the procedure quickly becomes intractable, since the size of the optimisation problem scales exponentially with the time horizon, number of uncertainties and uncertainty levels \cite{thangavel_robust_2017}.


Although many of the methods above focus on robustness, they do not incorporate uncertainty over the parameters of the transition function.
In \cite{bradford_stochastic_2017}, this is accounted for by using a SNMPC with an Unscented Kalman Filter to propagate the uncertainty over the state-space.
However, this method requires an optimisation with chance constraints to be solved online and, to keep the problem feasible, the variance of the trajectories has to be artificially constrained.
The most similar approach is perhaps presented in \cite{arruda_uncertainty_2017}, where the IT-MPC formulation is used in conjunction with a Ensemble of Mixture Density Networks (E-MDN) to approximate the joint probability distribution of states and actions.
This is similar to our approach, however as the E-MDN tries to approximate the joint distribution of states and actions, it needs to be retrained entirely on new environments.

In contrast, the variant of IT-MPC proposed in this paper uses the UT to propagate the uncertainty over model parameters.
This reduces the dimensionality of the inference problem and results in a controller more adept to generalise to unseen situations.
Moreover, unlike the stochastic optimisation strategies, our framework is very amenable to the inclusion of constraints, as the control update law is based on sampled trajectories.
As shown in \cite{williams_information-theoretic_2018} constraints may be applied directly to the control actions.
On the other hand, we can apply soft constraints to the state space through the cost function.
This is easily achieved as there is no need for the cost function to be differentiable and assures that a feasible solution will exist.

Additionally, DISCO takes advantage of the BayesSim Likelihood-free Inference (LFI) framework presented in \cite{ramos_bayessim:_2019} to update the model uncertainty periodically. 
Hence, given a set of true observations after a specified episode length, we can update our knowledge of the posterior probability density of parameters  $p(\simparams|\vec{x} = \vec{x}^r)$.
This way our model can adapt to variations in the environment, e.g.\ adjust friction coefficients in case of rain, or intrinsic to the transition function, e.g.\ change of weight distribution.
In contrast to other inference methods, such as Variational Inference or Markov Chain Monte Carlo, where a likelihood function is needed, in LFI we compute an approximated parametric distribution of the true posterior.
Furthermore, BayesSim was shown to be more data efficient than other LFI methods, such as Approximate Bayesian Computation \cite{ramos_bayessim:_2019}.

\section{Preliminaries}
\label{sec:prelim}

We consider the problem of controlling a discrete-time stochastic system described by a non-linear set of difference equations of the form:
\begin{align}
\label{eq:gen_problem}
\state_{t+1} &= f \left(\state_{t}, \stimuli_{t}\right)
\end{align}
where $f$ is the transition function, $\state_{t} \in \mathbb{R}^\stateDim$ denotes the system states, and $\stimuli_{t} \sim \normal \left( \control, \controlAuth \right) \in \mathbb{R}^\controlDim$ is the control input at a given time $t$.
We assume a finite time-horizon $\controlHorizon$, and that the control frequency is given.
Note that there is no direct control over the variable $\stimuli$, but we are able to control its mean $\control$.
This assumption considers not only a multiplicative noise model which is common in robotics, where lower-level actuator controllers are usually present, but also an amount of exploration in our control actions.
As such, in practice, $\controlAuth$ is a hyper-parameter of our control system that may need to be artificially increased.

More generally, we are interested in the problem where the real transition function $f \left( \mathbf{x, v} \right)$ is approximated by a parameterised non-linear forward model $f \left( \mathbf{x, v, \simparams} \right)$, represented as $f_\simparams$ for compactness.
Equation~(\ref{eq:gen_problem}) may then be rewritten as:
\begin{align}
\label{eq:gen_param_problem}
\state_{t+1} &= f_\simparams \left(\state_{t}, \stimuli_{t}\right).
\end{align}

\subsection{Information Theoretical MPC (IT-MPC)}

Following the steps in \cite{williams_information-theoretic_2018}, we can define a fixed length input sequence $U = \left( \control_0, \ldots, \control_{\controlHorizon-1} \right)$ over a fixed control horizon $\controlHorizon$, onto which we apply a Receding Horizon Control strategy.
This yields $\stimSeq = \left(\stimuli_0, \stimuli_1, \ldots, \stimuli_{\controlHorizon-1} \right) \in \mathbb{R}^{\controlDim \times \controlHorizon}$, which is itself a random variable.
Furthermore, let's denote as $\mathbb{P}$ the joint probability distribution and $p$ the corresponding probability density function (pdf) of the uncontrolled system (i.e. $U \equiv 0)$.
Likewise, $\mathbb{Q}$ is the joint distribution and $q$ the corresponding pdf for an open-loop control sequence.
The optimal control problem may then be be defined as:
\begin{align}
\label{eq:optimal_u_def}
\controlSeq^{*}=\argmin_{\controlSeq \in \controlSpace} \mathbb{E}_{\mathbb{Q}} \left[\phi\left(\state_{\controlHorizon}\right)+\sum_{t=0}^{\controlHorizon-1} \Loss\left(\state_{t}, \control_{t}\right)\right],
\end{align}
where $\controlSpace$ is the set of admissible controls, $\termCost\left(\state_{\controlHorizon}\right)$ is a terminal cost function, and $\Loss\left(\state_{t}, \control_{t}\right)$ is a running cost function of the form:
\begin{align}
\label{eq:loss}
\Loss\left(\state_{t}, \control_{t}\right)=\instCost\left(\state_{t}\right)+\frac{\invTemperature}{2}\left(\control_{t}^{\transpose} \controlAuth^{-1} \control_{t}+\controlMin_{t}^{\transpose} \control_{t}\right),
\end{align}
where $\invTemperature \in \mathbb{R}^{+}$ is known as the inverse temperature and the affine term $\controlMin$ allows the location of the minimum control (rest position) to be different from zero.
Noting that the state cost may be considered independent from the control terms, we can define $C\left(\state_{0}, \state_{1}, \ldots \state_{\controlHorizon}\right)=\phi\left(\state_{\controlHorizon}\right)+\sum_{t=0}^{\controlHorizon-1} \instCost\left(\state_{t}\right)$.
Moreover, we define a mapping operator, $\seqOp$, from input sequences $\stimSeq$ to their resulting trajectory by recursively applying $f_{\simparams}$ given $\state_0$, $\seqOp(\stimSeq; \state_0, \simparams) = \left[\state_0, f_{\simparams}(\state_0, \stimuli_0), f_{\simparams}(f_{\simparams}(\state_0, \stimuli_0), \stimuli_1), \ldots \right]$.
This leads to the following state cost function:
\begin{align}
    \stimCost\left(\stimSeq; \state_0, \simparams\right) = \stimCost\left(\stimSeq\right) = C\left(\seqOp \left(\stimSeq\right)\right).
\end{align}

Finally, IT-MPC relies on the \textit{free-energy} principle to compute a lower bound for the optimal control problem and defines the form of the optimal distribution function $q^*(\stimSeq)$ for which this bound is tight and achieves the optimal control $U^*$.
It can be shown that such distribution is of the form
\begin{align}
\label{eq:optimal_pdf}
q^{*}(\stimSeq) &=\frac{1}{\normaliser^{*}} \exp \left(-\frac{1}{\invTemperature} \stimCost(\stimSeq)\right) p(\stimSeq) \\ 
\normaliser^{*}     &=\int_{\mathbb{R}^{\controlDim \times \controlHorizon}} \exp \left(-\frac{1}{\invTemperature} \stimCost(\stimSeq)\right) p(\stimSeq) \mathrm{d} \stimSeq, 
\end{align}
where the base distribution $p(\stimSeq)$ has been augmented with the cost of the state trajectory.
This results in $\control_{i}^{*}=\mathbb{E}_{\mathbf{Q}^{*}}\left[\stimuli_{t}\right] \forall t \in\{0,1, \ldots \controlHorizon-1\}$.
Therefore, the optimal open-loop control sequence is the expected value of control trajectories sampled from the optimal distribution.
As we cannot sample directly from $\mathbb{Q}^{*}$, we can resort to importance sampling \cite{andrieu_introduction_2003} to construct an unbiased estimator of the optimal distribution, given the current control distribution, namely
\begin{align}
\mathbb{E}_{\mathbb{Q}} \left[\stimuli_{t}\right] =\int q^{*}(\stimSeq) \stimuli_{t} \mathrm{d} \stimSeq = \int \weight(\stimSeq) q(\stimSeq | \hat{U}, \controlAuth) \stimuli_{t} \mathrm{d}\stimSeq,
\end{align}
where $\weight(\stimSeq) = q^{*}(\stimSeq)/q(\stimSeq | \hat{U}, \controlAuth)$ is the importance sampling weight.
Therefore, we can switch the expectation to $\mathbb{E}_{\mathbb{Q}_{\hat{U}}}$, resulting in $\mathbb{E}_{\mathbb{Q}_{\hat{U}}}\left[\weight(\stimSeq) \stimuli_{t}\right]$.
We can then use the definition of the optimal distribution \wrt the base measure distribution given in \cite{williams_information-theoretic_2018} to derive the \textit{optimal information-theoretic control law}:
\begin{align}
\weight(\stimSeq)&=\frac{1}{\normaliser} \exp \left(-\frac{1}{\invTemperature}\left(\stimCost(\stimSeq)+\invTemperature \sum_{t=0}^{\controlHorizon-1} \control_t^\transpose \controlAuth^{-1} \stimuli_{t}\right)\right) \nonumber \\
\control_{t}&=\mathbb{E}_{\mathbf{Q}_{\hat{U}}}\left[\weight(\stimSeq) \stimuli_{t}\right],
\label{eq:control_law}
\end{align}
where:
\begin{equation}
\normaliser =\int \exp \left(-\frac{1}{\invTemperature}\left(\stimCost(\stimSeq)+\invTemperature \sum_{t=0}^{\controlHorizon-1}\control_{t}^\transpose \controlAuth^{-1} \stimuli_{t}\right)\right)~, \diff\stimSeq
\end{equation}
and $\control_{t}= \left(\hat{\control}_{t} - \tilde{\control}_{t}\right)$ is the difference between the current control action $\hat{\control}_{t}$ and the minimum control $\tilde{\control}_{t}$ (adjusted by $\controlMin$ and usually zero).
Note that in practice, for numerical stability, we multiply the numerator and denominator of $\weight(\stimSeq)$ by a factor $\exp \left(\frac{1}{\invTemperature} \rho\right)$, where $\rho$ is defined as the minimum cost.

\subsection{Likelihood-free parameter estimation}
\label{sec:lfi}
Recent advances in LFI allowed the use of probabilistic inference to learn distributions over simulation parameters \cite{ramos_bayessim:_2019}. The main idea is that of approximating an intractable posterior $p(\bm{\theta}|\mathbf{x}=\mathbf{x}^r)$ using data generated from a generative forward model (or simulator) where trajectories are collected for different simulation configurations. Therefore, one can directly learn a conditional density $q_\phi(\bm{\theta}|\mathbf{x})$ where parameters $\phi$ are learned through an optimisation procedure. The learned model usually takes the form of a mixture of Gaussians where inputs are summary statistics obtained from trajectories and outputs are the parameters of the mixture.

The goal is to maximise the likelihood $\prod_n q_\phi(\bm{\theta}_n|\mathbf{x}_n)$. It has been shown in previous work \cite{ramos_bayessim:_2019} that $q_\phi(\bm{\theta}|\mathbf{x})$ will be proportional to $\frac{\tilde{p}(\bm{\theta})}{p(\bm{\theta})}p(\bm{\theta}|\mathbf{x})$ if the log-likelihood is optimised as follows:
\begin{equation}
\label{eq:bayessim_loss}
{\cal L}(\phi)=\frac{1}{N}\sum_n\log q_\phi(\bm{\theta_n}|\mathbf{x}_n)    
\end{equation}
Consequently, a posterior estimate can be obtained by:
\begin{equation}
    \hat{p}(\bm{\theta}|\mathbf{x}=\mathbf{x}^r)\propto \frac{p(\bm{\theta})}{\tilde{p}(\bm{\theta})}q_\phi(\bm{\theta}|\mathbf{x}=\mathbf{x}^r).
    \label{eq:post}    
\end{equation}

The conditional density $q_\phi(\bm{\theta}|\mathbf{x})$ is a mixture of $K$ Gaussians,
\begin{equation}
\label{eq:mdn}
    q_\phi(\bm{\theta}|\mathbf{x})=\sum_{k=1}{K}\alpha_k(\mathbf{x}){\cal N}(\bm{\theta}|\bm{\mu(\mathbf{x})}_k,\bm{\Sigma}_k(\mathbf{x})),
\end{equation}
where $\{\alpha_k(\mathbf{x})\}_{k=1}^K$ are mixing functions, $\{\mu_k(\mathbf{x})\}_{k=1}^K$ are mean functions and $\{\Sigma_k(\mathbf{x})\}_{k=1}^K$ are covariance functions.

\section{DISCO}

At its core, model-based control relies on an approximated transition function to optimise the control actions over the control horizon.
In practice, this transition function is usually defined {\em a priori} using fundamental physical principles and domain knowledge, or empirically by applying system identification techniques \cite{simchowitz_learning_2018} or learning methods from data \cite{schaal_learning_1997, abbeel_autonomous_2010, simchowitz_learning_2018}.
Typically, these methods provide deterministic transition functions that do not incorporate model uncertainty and are invariant over time.
As discussed in \cite{williams_information_2017}, the closed-loop RHC offers a degree of robustness to model uncertainties, but the compounding error of poor predictions along the control horizon will reduce the stability margins of the system.
Using the methods outlined in \autoref{sec:prelim}, in this paper we propose a framework to apply the IT-MPC stochastic control formulation to problems where the parameters of the transition function $f_\simparams$ are unknown, but belong to a problem dependent \textit{prior}, $p(\simparams)$.
Furthermore, we make use of BayesSim \cite{ramos_bayessim:_2019} to refine our knowledge of the parameters as we interact with the environment and gather new observations.
The intuition behind this approach is that, by refining our knowledge of the parameters of an otherwise well-defined transition function, we will capitalise not only on the application domain knowledge, but also on the adaptability of inference-free learning methods.
Since $\simparams$ represents a plausible range of unknown physical parameters, e.g.\ mass or friction coefficient, it is straightforward to incorporate domain knowledge to this formulation.
Alternatively, an improper uninformative prior may be used when no assumptions are given.
On the other hand, by updating our knowledge of $p(\simparams | \state=\state^r)$ given observed data, we are more likely to cope with problems such as \textit{covariate shift} \cite{ganegedara_online_2016} and \textit{reality gap} \cite{ramos_bayessim:_2019, chebotar_closing_2018}.
The complete method is presented in \autoref{alg:disco}.

\subsection{Problem setup}

Given a forward model with parameters $\simparams$ and distributed according to $p(\simparams)$, trajectories can be obtained from it by first sampling $\simparams$ and generating roll outs by propagating the state-action pairs through the transition function. Although the parameters are stochastic, we assume they are invariant throughout the control horizon for a given trajectory.
This is a reasonable assumption as the latent parameters are usually stable physical quantities and the update frequency of the control loop is significantly faster than their time constants.
In this situation, the optimal distribution given in (\ref{eq:optimal_pdf}) becomes:
\begin{align}
\label{eq:disco_optimal_pdf}
q^{*}(\stimSeq, \simparams)=\frac{1}{\normaliser} \exp \left(-\frac{1}{\invTemperature} S_{\simparams}(\stimSeq)\right) p(\stimSeq|\simparams)p(\simparams),
\end{align}
where we overload the notation to emphasise the dependence of $\stimCost(\stimSeq)$ on the now stochastic $\simparams$.
However, as $\stimSeq$ and $\simparams$ are independent, we can drop the conditioning in $p(\stimSeq|\simparams)=p(\stimSeq)$.
As a result, our control law can be expressed as:
\begin{align}
\label{eq:disco_control_law}
\control_{t}=\mathbb{E}_{\mathbb{Q}_{\hat{U}}}\left[\mathbb{E}_{\mathbb{Q}_{\simparams}}\left[\omega_\simparams(\stimSeq) \stimuli_{t}\right]\right]=\mathbb{E}_{\mathbb{Q}_{\hat{U}, \simparams}}\left[w_\simparams(\stimSeq) \stimuli_{t}\right],
\end{align}
where $\omega_\simparams$ shows the dependence on $\simparams$ and we applied the law of total expectation to get the resulting equivalence.
This means our update rule now has to sample jointly from the distributions of $\stimSeq$ and $\simparams$.

\subsection{Propagation of uncertainty over the state-space dynamics}

If we sample sufficiently from $p(\stimSeq,\simparams)$, we are able to reconstruct the joint distribution $q(\stimSeq, \simparams)$ and compute our control updates.
However, we note that the increased dimensionality of the sample space requires the number of samples to grow combinatorially.
As such, we resort to the unscented transform \cite{julier_unscented_2004} as more efficient approach to propagate the uncertainty of $\simparams$ throughout the state-space.

In \cite{julier_unscented_2004}, the authors demonstrate how UT is able to reconstruct an approximate $Y^\prime(x)$ of the random variable $Y = g\left(X\right)$ resulting when an original random variable $X$ is applied to a non-linear function $g$.
The premise behind this approach is that it should be easier to approximate a probability distribution than a arbitrary non-linear transformation.
The idea is to select a set of \textit{sigma points} able to capture the most important statistical properties of the prior random variable $X$.
The necessary statistical information captured by the UT is the first and second order moments of $p(X)$.
The number of sigma-points needed to do this $\sigmaLength = 2\stateDim + 1$, where $\stateDim$ is the dimension of $X$.
In \cite{van_der_merwe_sigma-point_2004}, it is shown that matching the moments of $X$ up to the $\stateDim$th order implies matching the moments of $Y$ to the same order.
By using a larger number of sigma-points, skew and kurtosis can also be captured \cite{julier_scaled_2002}.

In DISCO, we refer to the formulation presented in \cite{van_der_merwe_sigma-point_2004} to compute sigma-points over the distribution $p(\simparams)$ of parameters. The expressions to compute the sigma-points and weights for the mean, $\sigmaWeight_{0}^{m}$, and covariance, $\sigmaWeight_{0}^{c}$, are presented below:
\begin{equation}
\label{eq:sigma_points}
\begin{split}
\begin{aligned}
\sigmaParam_{0}&=\mean{\simparams} & \quad
\sigmaParam_{i}&=\mean{\simparams}+(\sqrt{(\stateDim+\primScaling) \covMatrix_{\simparams})}_{1 \leq i \leq \stateDim} \\
\sigmaWeight_{0}^{m}&=\frac{\primScaling}{\stateDim+\primScaling} &
\sigmaParam_{i}&=\mean{\simparams}-(\sqrt{(\stateDim+\primScaling) \covMatrix_{\simparams})}_{\stateDim+1 \leq i \leq 2n}
\end{aligned}
\\
\begin{aligned}
\sigmaWeight_{0}^{c}=\sigmaWeight_{0}^{m}+\left(1-\primScaling^{2}+\degree\right) 
\quad
\sigmaWeight_{i}^{m}=\sigmaWeight_{i}^{c}=\frac{1}{2(\stateDim+\primScaling)},
\end{aligned}
\end{split}
\end{equation}
where $\primScaling = \spread^2(\stateDim+\secScaling) - \stateDim$ is the primary scaling factor, $\secScaling$ a secondary scaling (usually 0), $\spread$ determines the spread of the sigma points around $\mean{\theta}$, and $\degree$ is a scalar to provide an extra degree freedom.
The reader is encouraged to refer to \cite{van_der_merwe_sigma-point_2004} for details on hyperparameter selection.
The sigma-points are then applied recursively to the transition function $f_\simparams$ to compute the cost $S_\simparams \left(\stimSeq; \simparams = \sigmaParam_{i}\right)$ for $i \in \{1, \ldots, \sigmaLength\}$.
In practice, the sigma points on the state space are given by $\sigmaState=\seqOp(\sigmaParam)$.
To ensure the trajectory cost of each sigma-point can be summarised using the UT, it is necessary to apply the same action sequence $\stimSeq$ sampled \iid  to all points. 
Effectively, this means we need to replicate $\sigmaLength$ times each action sequence $\stimSeq$ during our update step.
Finally, the mean trajectory cost is given by
\begin{align}
    \stimCost(\stimSeq) = \sum_{i=0}^\sigmaLength \sigmaWeight_i^m \stimCost_\simparams(\stimSeq; \simparams = \sigmaState_i),
\end{align}
and used in (\ref{eq:control_law}) for the control law update.

\subsection{Updating the parameter prior distribution}

At each time step we are computing a new control action $\control_t$, applying it to our environment and collecting new observations $\state^r_{t+1}$.
The pairs of $[\control_t, \state^r_t]$ represent a trajectory $\trajectory$, up to a specified time-length.
This serves as input for the estimate of the posterior probability of $p(\simparams| \trajectory)$.
Once sufficient data has been aggregated in $\trajectory$, we can use the method presented in \cite{ramos_bayessim:_2019} to refine the posterior estimate.

Finally, the unscented transform requires as an input a mean vector $\mean{\simparams}$ and covariance matrix $\covMatrix_\simparams$ for the parameters.
Therefore, these have to be retrieved from $q_\phi(\simparams | \trajectory)$, or, alternatively, the highest weighted Gaussian may be selected if it is above a specified threshold.

\begin{algorithm}
\caption{DISCO}
\label{alg:disco}
\SetAlgoLined
\textbf{Control Hyperparameters}: $\invTemperature, \controlAuth, \controlMin, \instCost, \termCost$\;
\textbf{UT Hyperparameters}: $\primScaling, \secScaling, \spread, \degree$\;
\textbf{Given}:
$f_\simparams$,
$p(\simparams)$,
$\controlSeq_0$,
$\controlHorizon$,
$\controlSamples$,
$\sigmaLength$,
$\trajectory$\;
\BlankLine
\emph{Update posterior distribution}\;
$q_{\phi}\left(\simparams | \trajectory\right) \gets \text{BayesSim}(\trajectory)$\;
$p\left(\simparams\right) \leftarrow q_{\phi}\left(\simparams | \trajectory\right)$\;
\While{task not complete}{
    $\state_0 \gets$ GetStateEstimate()\;
    \For{$k\gets0$ \KwTo $\controlSamples-1$}{
        Sample $\mat{\mathcal{E}}^k=\left(\vec{\epsilon}_0^k \ldots \vec{\epsilon}_{\controlHorizon-1}^k\right), \vec{\epsilon}_t^k\sim\normal(0, \controlAuth)$\;
        \For{$i\gets1$ \KwTo $\sigmaLength$}{
            $\simparams_i \gets \simparams \sim p(\simparams)$ (MC) or $\sigmaParam_i$ (UT)\; 
            $\state\gets\state_0$\;
            \For{$t\gets1$ \KwTo $\controlHorizon$}{
                $\state\gets f_\simparams(\state, \stimuli_t, \simparams_i)$\;
                $\stimCost^k_i \pluseq \instCost(\state) + \invTemperature\control_{t-1}^\transpose \controlAuth^{-1} \left(\vec{\epsilon}_{t-1}\right)$\;
            }
            $\stimCost^k_i \pluseq \termCost(\state)$\;
        }
        $\stimCost^k = \sum_{i=1}^L \sigmaWeight^m_i \stimCost^k_i$\;
    }
    $\rho \gets \min(\stimCost^k)$\;
    $\normaliser \gets \sum^\controlSamples_{k=1} \exp \left(-\frac{1}{\invTemperature}\left(\stimCost^{k}-\rho\right)\right)$\;
    \For{$k\gets1$ \KwTo \controlSamples}{
        $\weight_{k} \gets \frac{1}{\normaliser} \exp \left(-\frac{1}{\invTemperature}\left(\stimCost^{k}-\rho\right)\right)$\;
    }
    \For{$t\gets0$ \KwTo $\controlHorizon-1$}{
        $\control_t \pluseq \sum_{k=0}^{\controlSamples-1} \weight\left(\mat{\mathcal{E}}^k \right)\vec{\epsilon}^{k}_{t}$\;
    }   
    SendToActuators$(\control_0)$\;
    Append$(\trajectory,[\state_0, \control_0])$\; 
    RollControlActions$(\control)$\;
}
\end{algorithm}

\section{Experimental results}

\subsection{Inverted pendulum swing-up task}

In this task, the controller has to swing and hold a pendulum upright using a torque command applied directly to the joint of a rigid-arm.
We used the simulator in \cite{brockman_openai_2016}, and always set the pendulum initial state to the downright position and at rest.
The state cost function used was $\instCost = 50\cos(\theta-1)^2 + \dot{\theta}^2$, and the terminal cost function $\termCost$ was set to zero.
The inverse control temperature $\invTemperature$ was set at 10 and the control authority $\controlAuth$ at 1.
We have also defined the number of sampled trajectories $\controlSamples=500$ and the control horizon $\controlHorizon=30$.
For more details on the experiment parameters, please refer to the \hyperref[sec:appendix]{Appendix}.

\begin{figure}
    \centering
    \epsfysize=60mm
    \epsfbox{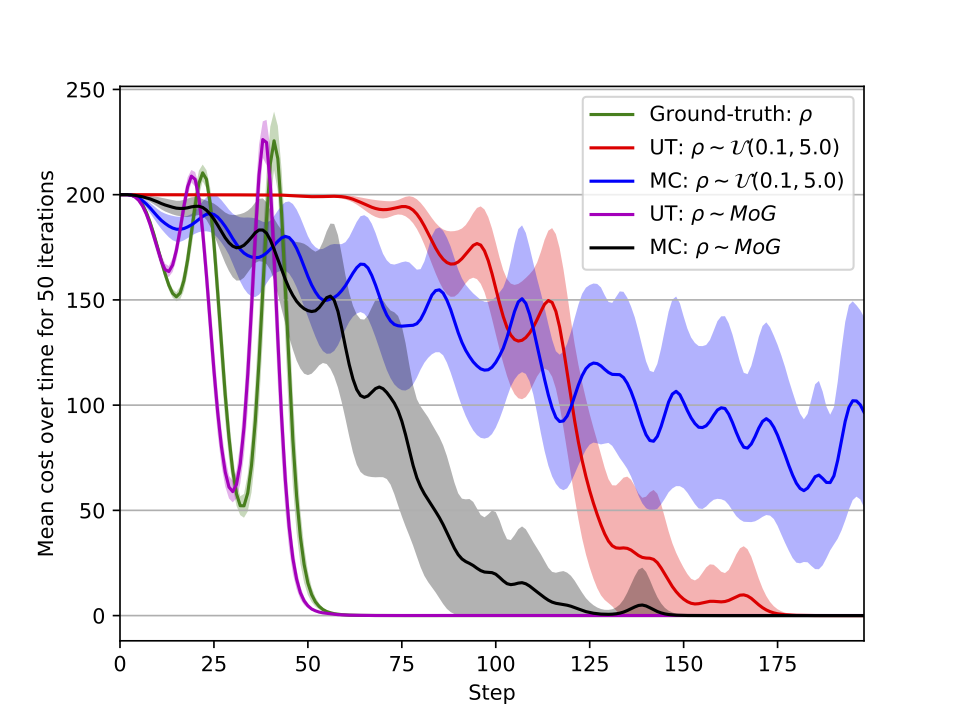}
    \caption{
    Mean cost over time for the inverted pendulum experiment. Shaded area represents one standard deviation.
    Three models where evaluated: a standard IT-MPC with access to the true system parameters (in green); DISCO using unscented transform with a prior distribution over parameters (in red) and with an updated posterior distribution (in magenta); and DISCO using MC sampling with a prior distribution (in blue) and an updated posterior (in black).
    }
    \label{fig:pendulum_cost}
\end{figure}

The results presented in \autoref{fig:pendulum_cost} are the mean cost over time for 50 iterations, for a baseline case, DISCO, and Monte Carlo sampling.
Note that the oscillatory behaviour of the cost function is expected, as the controller has insufficient authority to balance the pendulum without the swinging action to increase the momentum.

All models used the same hyperparameters described above, with the exception of the $\controlHorizon$ for the case of MC sampling.
Given that we want to compare the performance of the controller when using UT against MC, for the case where MC is used, we increment the amount of trajectories sampled by the number of sigma points $\sigmaLength$ used by the UT.
Effectively, for the MC controller we have $\controlSamples=2500$ trajectories.

The unknown parameters in this example were the length of the arm and the mass of the pendulum.
As a prior, we assumed an uniform distribution between $0.1$ and $5$ for both parameters.
The posterior distribution was given by a mixture of Gaussians with $5$ components, trained using a reference control policy.
Note that in our simulations, both models shared the same posterior distribution estimate.
Once trained and conditioned on the observed data, the resulting mixture had a mean estimate for the length of $0.89$ meter and for mass $0.90$ kilo.
The covariance matrix was diagonal, and the variance was $0.01$ for the length estimate and $0.03$ for mass.
One of the components of the mixture was dominant with a weight of $0.979$ and was used as reference for the UT.

DISCO with UT outperforms MC sampling both with an uninformative prior and inferred posterior.
Noticeable also, the performance of UT with the posterior distribution is better than the baseline model.
This is explainable by the fact that the parameter randomisation introduced by the sigma-points provides more information in the trajectory evaluation.
This way, trajectories that are borderline to a higher cost state captured by one of the sigma points get penalised.
Effectively, UT works like an automatic calibration of the control temperature, when the prior is broad, many trajectories are considered in the control update average.
Conversely, when the posterior gets refined, the controller is more confident to select fewer trajectories.

\subsection{Skid-steer robot}
\begin{figure}
    \centering
    \subfloat[Robot]{\includegraphics[width=0.5\columnwidth]{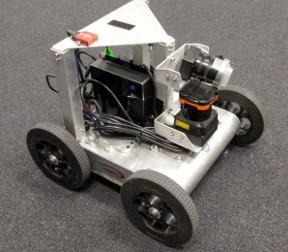}\label{fig:wombot-robot}}
    \subfloat[Marginals]{\includegraphics[width=0.5\columnwidth]{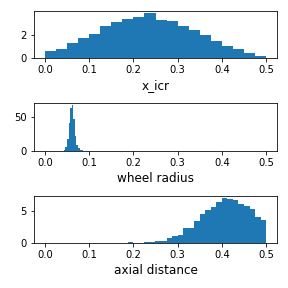}\label{fig:wombot-marginals}}
    \caption{Skid-steer robot and its parameter estimates}
    \label{fig:wombot}
\end{figure}

This section presents experimental results with a physical robot equipped with a skid-steering drive mechanism (\autoref{fig:wombot}).
We modelled the kinematics of the robot based on a modified unicycle model, which accounts for skidding via an additional parameter \cite{kozlowski_modeling_2004}.
The parameters to be estimated via BayesSim are the robot's wheel radius $\wradius$, axial distance $\axdist$, i.e. the distance between the wheels, and the displacement of the robot's instant centre of rotation (ICR) from the robot's centre $\xicr$. A non-zero value on the latter affects turning by sliding the robot sideways. To estimate the parameters, the robot was driven manually around a circle and had its trajectory data recorded.
From the trajectory data we computed cross-correlation summary statistics as $(\mean{x},\mean{y},\mean{\Delta x},\mean{\Delta y})$, which capture the centre of trajectory and the average linear velocity.
In simulation, the wheel speed commands sent to the robot were repeated $N=1000$ for different parameter settings sampled from a uniform prior, $\xicr \sim [0,0.5]$, $\wradius\sim[0,0.5]$, $\axdist \sim [0.1,0.5]$.

\autoref{fig:wombot-marginals} presents the resulting marginal estimates from BayesSim for each parameter of the robot's kinematic model.
For comparisons, physical measurements indicate a $\wradius$ of around 0.06 m and $\axdist$ of around 0.31 m.
Measuring $\xicr$, however, involves a laborious process, which would require different weight measurements or many trajectories from the physical hardware \cite{yi_kinematic_2009}.
As we are only applying a relatively simple kinematic model of the robot to explain the real trajectories, the effects of the dynamics and ground-wheel interactions are not accounted for.
As a result, BayesSim tries to compensate for the miss-specifications in some parameters estimation, such as the axial distance.
This explains the larger variation in $\axdist$, and consequently $\xicr$.

\begin{figure}
    \centering
    \includegraphics[width=\columnwidth]{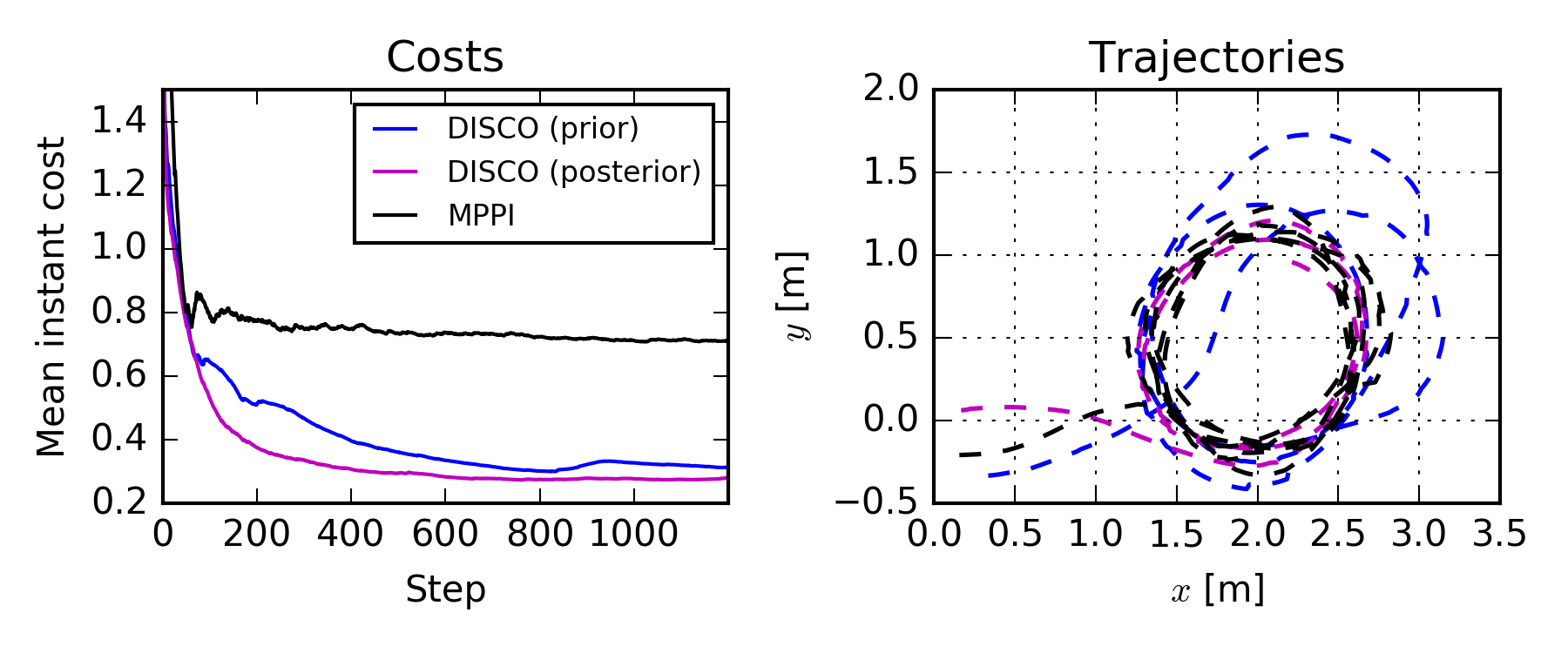}
    \caption{Results with physical robot}
    \label{fig:wombot-results}
\end{figure}

The control task was defined as following a circular path at a constant tangential speed.
Costs were set to make the robot follow a circle of 0.75 m radius with $c(\state_t) = \sqrt{d_t^2+(s_t-s_0)^2}$, where $d_t$ represents the robot's distance to the edge of the circle and $s_0=0.2$ m/s is a reference linear speed.
We performed experiments sampling from the uniform prior over the parameters $p(\simparams)$, sampling from the posterior $q(\simparams|\trajectory)$, and using only a point estimate set to $\xicr=0.12$, $\wradius=0.06$ and $\axdist=0.47$, which was adjusted offline to reduce simulation error.
For clarity, instead of the noisy raw costs, we present the mean instant cost, i.e. $\mean{\instCost}_t=\frac{1}{t}\sum_{i=1}^t\instCost(\state_i)$ and the executed trajectories in \autoref{fig:wombot-results}. For the complete experiment parameters refer to the \hyperref[sec:appendix]{Appendix}.
 
We see that considering parameter uncertainty via DISCO provides significant performance improvements over the baseline IT-MPC algorithm running with a point estimate.
Although, in term of costs, both the prior and posterior estimates offer similar performance, we see the advantages of using the parameter posterior estimates in the trajectories plot, where we see overshooting happening on some portions of the path.
The latter can be explained by the prior allowing kinematic parameters candidates that are too far from the true values. Additionally, refining the posterior distribution allows the system to adapt to new configurations or drift in the model parameters.
Lastly, a noisier speed control explains the gap between the baseline MPPI and the DISCO methods, despite the similar performances in terms of path tracking.



\section{Conclusion}

This paper is a first step towards incorporating model uncertainty and sophisticated Bayesian inference methods to stochastic model based control.
We showed how uncertainty over parameters may be formally incorporated into an stochastic non-linear MPC controller and evaluated methods of propagating the uncertainty into trajectory roll outs.
This extension to information theoretical MPC provides the building blocks of an adaptive controller framework, more resilient to issues arising from reality gap and covariate shift.
As shown in the robotic experiments, incorporating uncertainty may lead to a more accurate assessment of the environment and increase the performance.

The unscented transform proved an efficient way to propagate uncertainty, reducing the burden of sampled trajectories.
When combined with the ability to impose hard-penalties on the state cost, the result is similar to a chance constraint, where the resulting trajectories from sigma points that violate the soft constraints are heavily penalised.
It is worth noticing, that this deterministic method of estimating the moments of the parameter distribution allow the task of sampling actions to be parallelised asynchronously and aggregated when computing the final cost. 
In future work we intend to explore further possibilities of uncertainty propagation in a principled way.

More importantly, we showed how LFI is a powerful tool to refine the estimation of the posterior distribution.
As the inference is based on the same transition function $f_\simparams$ of the controller, it may compensate overly simplified models of the environment.
Therefore, we want to explore pathways to efficiently retrain this estimate online so practical experiments with time-variant parameters may be conducted.
This is a crucial step towards generalisation of control policies for autonomous robots operating under varying environments and configurations.
Crucially, by combining parameter estimation and gradient-free control methods, DISCO may also be used with black-box simulators, such as data-driven function approximators, as long as we are able to sample efficiently from them.
This is a promising direction for future research.

\addtolength{\textheight}{-7cm}   

\bibliographystyle{IEEEtran}
\bibliography{references}

\addtolength{\textheight}{7cm}   
\newpage
\onecolumn
\appendix[Parameters used in the experimental results]
\label{sec:appendix}
A comprehensive list of parameters used in the experimental section are listed on \autoref{tab:pendulum_param} and \autoref{tab:skidsteer_param}.
For both experiments, the unscented transform secondary scaling ($\secScaling$) and minimum control ($\controlMin$) were set to zero. Note that, as the random seeds were not controlled, slight variations are expected when reproducing the results. Similarly, the update of the posterior distribution approximation, $q_\phi(\simparams | \trajectory)$, will depend on $\trajectory$ and therefore will vary in every execution.

\begin{table}[htb!]
\centering
\caption{Parameters for the inverted pendulum experiment.}
\label{tab:pendulum_param}
\begin{tabular}{@{}ll@{}}
\textbf{Parameter} & \textbf{Inverted Pendulum} \\ \midrule
Sampled actions ($\controlSamples$) & $500$ \\
Control horizon ($\controlHorizon$) & $30$ \\
Inverse temperature ($\invTemperature$) & $10$ \\
Control authority ($\controlAuth$) & $1$ \\
Instant state cost ($\instCost$) & $50\cos(\theta-1)^2 + \dot{\theta}^2$ \\
Terminal state cost ($\termCost$) & $0$ \\
Sigma points ($\sigmaLength$) & $5$ \\
UT Spread ($\spread$) & $0.5$ \\
UT scalar ($\degree$) & $2$ \\
Prior distribution ($p(\simparams)$) & \\
- over pole length $l$ & $\mathcal{U}(0.1, 5)$ \\
- over pole mass $m$ & $\mathcal{U}(0.1, 5)$ \\
Posterior distribution ($q_\phi(\simparams | \trajectory)$) & \\
- over pole length $l$ & $\mathcal{N}(0.89, 0.01)$ \\
- over pole mass $m$ & $\mathcal{N}(0.9, 0.03)$ \\
\end{tabular}
\end{table}

\begin{table}[htb!]
\centering
\caption{Parameters for the skid-steer experiment.}
\label{tab:skidsteer_param}
\begin{tabular}{@{}ll@{}}
\textbf{Parameter} & \textbf{Skid-steer Robot} \\ \midrule
Sampled actions ($\controlSamples$) & $400$ \\
Control horizon ($\controlHorizon$) & $50$ \\
Inverse temperature ($\invTemperature$) & $0.1$ \\
Control authority ($\controlAuth$) & $0.25$ \\
Instant state cost ($\instCost$) & $c(\state_t) = \sqrt{d_t^2+(s_t-s_0)^2}$ \\
Terminal state cost ($\termCost$) & $0$ \\
Sigma points ($\sigmaLength$) & $7$ \\
UT Spread ($\spread$) & $0.5$ \\
UT scalar ($\degree$) & $2$ \\
Prior distribution ($p(\simparams)$) & \\
- over $\xicr$ & $\mathcal{U}(0, 0.5)$ \\
- over $\wradius$ & $\mathcal{U}(0, 0.5)$ \\
- over $\axdist$ & $\mathcal{U}(0.1, 0.5)$ \\
Posterior distribution ($q_\phi(\simparams | \trajectory)$) & \\
- $\left[\xicr, \wradius, \axdist\right]^\transpose$  & $\mathcal{N}(\mu,\Sigma)$ \\
- $\mu$  & \,$\left[\begin{array}{ccc}0.238 & 0.061 & 0.415\end{array}\right]^\transpose$ \\
- $\Sigma \times 10^{-3}$  & 
$
\left[
\begin{array}{ccc}
0.13 & -0.03 & -0.04\\
-0.03 & 0.15 & 0.03\\
-0.04 & 0.03 & 0.09\\
\end{array}
\right]
$ \\
\end{tabular}
\end{table}

\end{document}